\documentclass[10pt,twocolumn,letterpaper]{article}

\usepackage{cvpr}       

\usepackage{graphicx}
\usepackage{amsmath}
\usepackage{array}
\usepackage{amssymb}
\usepackage{booktabs}
\usepackage{multirow}
\usepackage[table,xcdraw]{xcolor}
\usepackage[accsupp]{axessibility}

\usepackage[pagebackref,breaklinks,colorlinks]{hyperref}

\usepackage[capitalize]{cleveref}
\usepackage{rotating}
\crefname{section}{Sec.}{Secs.}
\Crefname{section}{Section}{Sections}
\Crefname{table}{Table}{Tables}
\crefname{table}{Tab.}{Tabs.}

\def\cvprPaperID{7237} 
\def\confName{CVPR}
\def\confYear{2022}

\begin{document}

\title{GaTector: A Unified Framework for Gaze Object Prediction}
\author{Binglu\ Wang\thanks{Corresponding author.}, \ \ \ \ Tao\ Hu, \ \ \ \ Baoshan\ Li, \ \ \ \ Xiaojuan\ Chen,\ \ \ \ Zhijie\ Zhang\\Xi’an University of Architecture and Technology, \ \ China}
\maketitle{}

\begin{abstract}
Gaze object prediction is a newly proposed task that aims to discover the objects being stared at by humans. It is of great application significance but still lacks a unified solution framework. An intuitive solution is to incorporate an object detection branch into an existing gaze prediction method. However, previous gaze prediction methods usually use two different networks to extract features from scene image and head image, which would lead to heavy network architecture and prevent each branch from joint optimization. In this paper, we build a novel framework named GaTector to tackle the gaze object prediction problem in a unified way. Particularly, a specific-general-specific (SGS) feature extractor is firstly proposed to utilize a shared backbone to extract general features for both scene and head images. To better consider the specificity of inputs and tasks, SGS introduces two input-specific blocks before the shared backbone and three task-specific blocks after the shared backbone. Specifically, a novel \textit{Defocus} layer is designed to generate object-specific features for object detection task without losing information or requiring extra computations. Moreover, the energy aggregation loss is introduced to guide the gaze heatmap to concentrate on the stared box. In the end, we propose a novel wUoC metric that can reveal the difference between boxes even when they share no overlapping area. Extensive experiments on the GOO dataset verify the superiority of our method in all three tracks, \ie object detection, gaze estimation, and gaze object prediction.

\end{abstract}

\section{Introduction}

\begin{figure}[thbp]
\centering
\setlength{\abovecaptionskip}{0.1cm}
\includegraphics[width=1\linewidth]{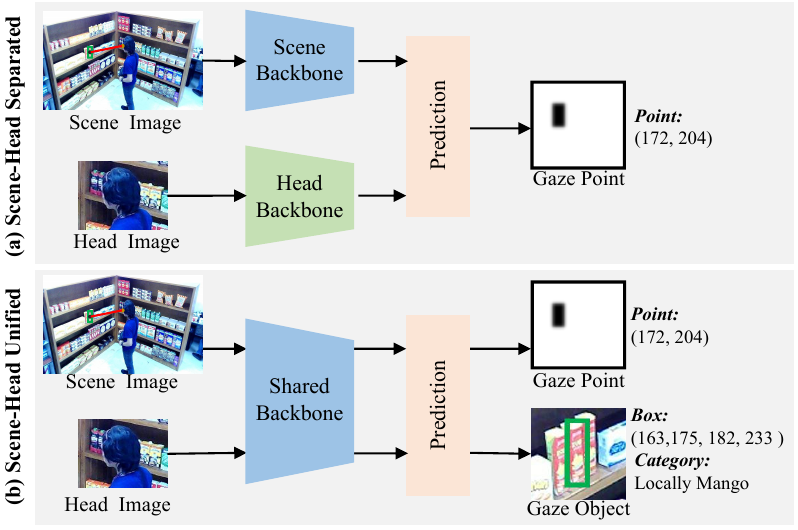}
\caption{Illustration of the gaze object prediction approaches. (a) Previous gaze prediction methods use two separate backbone to tackle scene image and head image, respectively. (b) The proposed SGS feature extractor can produce features from scene image and head image in a unified manner.}
\vspace{-0.6cm}
\label{Illustration}
\end{figure}

Gaze estimation (GE) aims at determining the direction and point that a person is staring at \cite{recasens2015they}. As gaze behavior is an essential aspect of human social behavior\cite{emery2000eyes,land2009looking} so we can infer potential information based on the staring object. For example, in front of the bus station, a person looking at his watch may indicate that he has something urgent to do. In shopping malls, customers staring at the product may want to purchase it. The stared object can generally reveal our state, \eg what they are doing, or what they plan to do.

In the gaze estimation community, researchers usually employ two separated backbones to process the entire scene image and the head image.  This scene-head separated structure was first proposed by Recasens \etal~\cite{recasens2015they}, where one backbone captures holistic cues from the entire scene image and the other backbone analyzes the details from the head image. Later, Lian~\etal \cite{lian2018believe} proposed the multi-scale gaze direction fields to precisely analyze the head image. Recently, Chong \etal\cite{chong2020detecting} elaborately designed interactions between the head branch and the scene branch and utilized deconvolutional layers to produce a fine-grained heatmap. In general, the gaze estimation performance keeps improving in recent years while the network architecture gradually gets complex. Existing models only predict the gaze area that people may stare at rather than precisely predicting the location of the object been be stared at. Tomas \etal\cite{tomas2021goo} recently pointed out that it has significant practical usage to identify the stared thing. As shown in Fig.~\ref{Illustration}, from the bounding box of the stared item, we can infer that the person is likely to buy the \textit{Locally Mango} product, which is beyond the scope of the traditional gaze estimation task.

However, performing gaze object prediction is non-trivial and faces the following challenges, \ie heavy network architecture, inconsistent requirements about image size. First of all, although it is an intuitive solution that adds an additional object detection branch to the existing two-branch gaze estimation models \cite{guan2020enhanced, zhao2020learning, saran2018human, recasens2017following, lian2018believe, chong2020detecting, recasens2015they,liu2021goal}, this approach would undoubtedly increase the number of calculations and parameters of the entire network. Besides, gaze estimation models usually employ an image of ordinary size (\eg $224 \times 224$) to capture a global receptive field. In contrast, objects in the retail scenario are generally small and dense, requiring an enlarged image to detect bounding boxes precisely. Moreover, compared with individuals carrying out gaze estimation and object detection, a more suitable way should employ a unified framework and achieve joint optimization.

To alleviate the above issues, we make the following three designs. (1) Different from previous gaze estimation works~\cite{recasens2015they, lian2018believe, chong2020detecting} that use two independent branches (see in Fig.~\ref{Illustration} (a)), we propose a specific-general-specific (SGS) mechanism to extract task-specific features from the scene and head image with only one backbone (see in Fig.~\ref{Illustration} (b)), which can help to reduce the parameters and computational burdens, and also make it possible to joint optimization for different inputs.
(2) To assist precise object detection, we develop a \textit{Defocus} layer to generate object-specific features. In particular, the input image with ordinary size can not produce feature map with sufficient resolution for detecting small and dense retail objects. The proposed \textit{Defocus} layer zooms the spatial size via shrinking the channel size, which can produce feature of high-resolution without losing information nor brings extra computations.
(3) To tackle the performance bottleneck of imprecise gaze heatmap, we propose the energy aggregation loss that measures the percentage of energy within the stared box, and use the ground truth bounding box to guide the gaze estimation process.

\begin{figure*}[thbp]
\centering
\setlength{\abovecaptionskip}{0.1cm}
\includegraphics[width=0.9\linewidth]{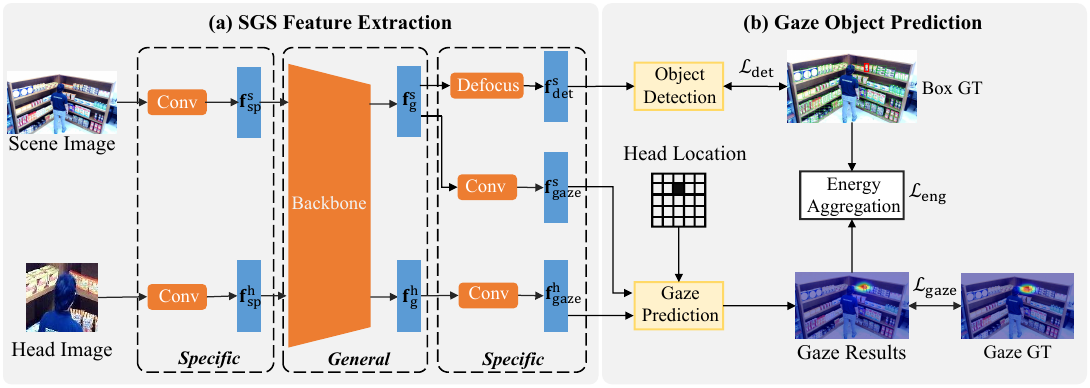}
\caption{Overview of the proposed method. (a) The specific-general-specific (SGS) architecture can provide task-specific features while sharing backbone. (b) The object detection module and gaze prediction module can be jointly trained by the energy aggregation loss.}
\vspace{-0.5cm}
\label{figGOPFramework}
\end{figure*}

In our work, we propose a unified framework, namely GaTector, to estimate gaze heatmap, detect retail objects and conduct gaze object detection, as shown in Figure \ref{figGOPFramework}. The scene and head images are first jointly tackled by the SGS feature extractor. Then, the object detection head discovers bounding boxes, and the gaze prediction head predicts gaze heatmap, so we can jointly consider gaze prediction and object detection results to carry out gaze object prediction. Also, a novel wUoC metrics proposed to better reveal the difference between boxes even when they share no overlapping area. Our contributions can be summarized as follows: 
\begin{itemize}
    \vspace{-0.2cm}
    \item We proposed a unified method GaTector with a novel wUoC evaluation metric to make an early exploration on the gaze object detection task.
    \vspace{-0.2cm}
    \item We propose a novel SGS mechanism that can extract task-specific features by a single backbone while maintaining satisfactory performance. A \textit{Defocus} layer is introduced in SGS to prepare high-resolution feature maps for small retail object detection, and the energy aggregation loss guides the gaze heatmap to be concentrated.
    \vspace{-0.2cm}
    \item On the large-scale GOO dataset, we consistently improve the performance on two traditional tracks, \ie gaze estimation, and object detection, while reducing the model parameter and computational costs. In addition, we build a solid baseline for the gaze object detection task to promote future research.
\end{itemize}

\vspace{-0.2cm}
\section{Related Work}
\vspace{-0.2cm}

\noindent
\textbf{Gaze following.} As a practical technique\cite{wang2022contextual, wang2022detail}, the gaze following task is proposed by Recasens \etal~\cite{recasens2015they} and is a well-explored branch of gaze estimation. Gaze estimation reveals where a person is looking and serves as an essential clue for understanding human intention\cite{wang2021multiple, wang2021exploring}. Existing gaze estimation works can be divided into three categories according to different scenarios \ie, gaze point estimation\cite{liu2021goal,liu2021generalizing}, gaze following\cite{triesch2006gaze,brooks2005development,li2021looking}, and 3D gaze estimation\cite{masse2019extended, elmadjian20183d}.

This paper is related to the gaze following task. In the early phase, Zhu \etal~\cite{zhu2012face} proposed a unified model that can perform face detection, pose estimation, and landmark estimation in realistic images. Recasens \etal\cite{recasens2015they} predict the gaze area by extracting head pose and gaze direction via deep model. Afterward, Parks \etal\cite{parks2015augmented} combined saliency maps with human head pose and gaze direction to predict the gaze area for the observer. In addition, there are also some works focused on general gaze following \cite{mukherjee2015deep,triesch2006gaze,brooks2005development}. For example, Mukherjee~\etal \cite{mukherjee2015deep} restored the interaction with the environment based on head pose estimation. Different from predicting gaze area, this paper studies the gaze object prediction task and aims at discovering the bounding box of the stared objects, which is more challenging.

\noindent
\textbf{Object detection.} Recently, object segmentation \cite{zhou2020matnet, zhou2022group, huang2021clrnet, huang2021scribble} and detection methods has received obvious progress, where detection methods can be divided into anchor-free detectors~\cite{redmon2016you,redmon2017yolo9000,redmon2018yolov3,bochkovskiy2020yolov4,tian2019fcos} and anchor-based detectors~\cite{yao2020automatic, ren2015faster, feng2020progressive, lin2017focal,feng2020tcanet}. As a representative method for the anchor-free paradigm, YOLO~\cite{redmon2016you} predicts bounding boxes at points close to the object's center. Later, a series of methods \cite{redmon2017yolo9000,redmon2018yolov3,bochkovskiy2020yolov4} based on YOLO gradually improved object detection performance and becoms an effective solution to object detection. The anchor-based methods include two categories: one-stage methods \cite{liu2016ssd,lin2017focal} and two-stage methods \cite{ren2015faster,he2017mask}. The one-stage detector can directly make predictions based on the extracted feature map and default anchors, while the two-stage detector first generate object proposals and then performs detailed refinement.

\noindent
\textbf{Gaze prediction in the retail industry.} In recent years, the automatic retail system and human-object interaction \cite{zhou2021cascaded, zhou2020cascaded} has aroused increasing research interests. For example, Harwood \etal~\cite{harwood2014mobile} use mobile phones to track the visual attention of consumers in the retail environment. Besides,new benchmarks are proposed to serve the automatic checkout systems, \eg the D2S dataset~\cite{follmann2018mvtec} and the RPC dataset~\cite{wei2019rpc}. Recently, EyeShopper~\cite{bermejo2020eyeshopper} offered an innovative system to achieve a precise estimation of customers' sight. In the retail environment, it is valuable for estimating the stared products to perform exact recommendations. However, existing methods only predict gaze area and leave the gaze object prediction problem under-explored.

\section{Method}

Given a scene image $\mathbf{I}_{\rm s}$ and head location mask $\mathbf{H}$, the head image $\mathbf{I}_{\rm h}$ usually is generated by cropping the scene image $\mathbf{I}_{\rm s}$. The goal of gaze object prediction task is to predict the bounding box and category label of the object started by a human. 

\subsection{SGS Mechanism}
To extract holistic scene features and detailed head features, traditional gaze follow works \cite{recasens2015they, lian2018believe, chong2020detecting} usually use two independent networks to tackle scene image $\mathbf{I}_{\rm s}$ and head image $\mathbf{I}_{\rm h}$, respectively. However, if this paradigm was used to resolve the gaze object detection problem, an extra object detection branch is needed and makes there exist three parallel branches in the model, and results in a significant increase of the parameters and computational cost. An intuitive way to reduce the parameters and computational cost is to share the backbone when extracting features for the object detection branch and gaze prediction branch. However, this manner will significantly reduce the performance (see in Table~\ref{tabCmpGaze}) as the information required by the object detection and gaze prediction are different. To alleviate this issue, this paper proposes a specific-general-specific (SGS) mechanism to jointly consider the task specificity of scene image and head image before and after the shared backbone. SGS employs a sharing backbone to extract features in a general manner and utilizes specific heads to prepare task-specific features for gaze prediction and object detection.

As shown in Fig.~\ref{figGOPFramework} (a), we first use two independent convolutional layers to extract input-specific features for scene image and head image before the shared backbone:
\begin{equation}
  \mathbf{f}_{\rm sp}^{\rm s} = \psi^{\rm s} (\mathbf{I}_{\rm s}), \ \ \mathbf{f}_{\rm sp}^{\rm h} = \psi^{\rm h} (\mathbf{I}_{\rm h}),
\end{equation}
where $\psi^{\rm s}(\cdot)$ and $\psi^{\rm h}(\cdot)$ denotes convolutional layers for scene image and head image, respectively. $\mathbf{f}_{\rm sp}^{\rm s}$ and $\mathbf{f}_{\rm sp}^{\rm h}$ denotes the extracted features. 
Then we feed $\mathbf{f}_{\rm sp}^{\rm s}$ and $\mathbf{f}_{\rm sp}^{\rm h}$ into the shared backbone and produce features in a general manner:
\begin{equation}
  \mathbf{f}_{\rm g}^{\rm s} = \psi^{\rm b} (\mathbf{f}_{\rm sp}^{\rm s}), \ \ \mathbf{f}_{\rm g}^{\rm h} = \psi^{\rm b} (\mathbf{f}_{\rm sp}^{\rm h})
\end{equation}
where $\mathbf{f}_{\rm g}^{\rm s}$ and $\mathbf{f}_{\rm g}^{\rm s}$ denotes the general features for scene image and head image, $\psi^{\rm b}$ denotes the shared backbone. By sharing the backbone, the network parameters and computational cost would be greatly reduced. Afterward, we take the general scene feature $\mathbf{f}_{\rm g}^{\rm s}$ as input and design a novel \textit{Defocus} layer (see in Section~\ref{secObjectDetection}) to generate object-specific feature for detecting the location:
\begin{equation}
  \mathbf{f}_{\rm det}^{\rm s} = \phi^{\rm det} (\mathbf{f}_{\rm g}^{\rm s})
\end{equation}
where $\phi^{\rm det}$ denotes the proposed \textit{Defocus} layer, $\mathbf{f}_{\rm det}^{\rm s}$ denotes the feature for object detection. Then, both general head feature $\mathbf{f}_{\rm g}^{\rm h}$ and general scene feature $\mathbf{f}_{\rm g}^{\rm s}$ are sent into two independent convolution layers to generate gaze-specific features for predicting gaze results:
\begin{equation}
  \mathbf{f}_{\rm gaze}^{\rm s} = \phi^{\rm s} (\mathbf{f}_{\rm g}^{\rm s}), \ \ \mathbf{f}_{\rm gaze}^{\rm h} = \phi^{\rm h} (\mathbf{f}_{\rm g}^{\rm h}),
\end{equation}
where $\phi^{\rm s}(\cdot)$ and $\phi^{\rm h}(\cdot)$ indicate two independent convolution layers. $\mathbf{f}_{\rm gaze}^{\rm s}$ and $\mathbf{f}_{\rm gaze}^{\rm h}$ indicate the scene features and head features for the gaze prediction network (Section~\ref{sectionGazePrediction}).

By sharing the backbone and generating specific features before and after the backbone, the proposed SGS mechanism can significantly reduce the parameters and computational cost while maintaining a satisfactory performance, forming an effective and efficient unified framework for gaze object detection.

\subsection{Gaze prediction}
\label{sectionGazePrediction}

\begin{figure}[!t]
\centering
\setlength{\abovecaptionskip}{0.1cm}
\includegraphics[width=0.9\linewidth]{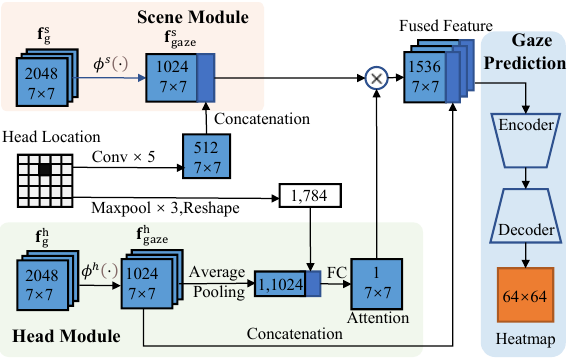}
\caption{Illustration of the gaze prediction network.}
\vspace{-0.4cm}
\label{figGazePrediction}
\end{figure}

As shown in Fig.~\ref{figGazePrediction}, the gaze prediction network takes two gaze-specific features (\ie $\mathbf{f}_{\rm gaze}^{\rm s}$ and $\mathbf{f}_{\rm gaze}^{\rm h}$) and the head location map (\ie $\mathbf{H}$) as input to predict the gaze results.

As the head location map can provide valuable guidance for the gaze prediction task, Chong \etal \cite{chong2020detecting} concatenate the head location map and scene image before extracting holistic scene features. However, backbone in our GaTector serves both the gaze prediction task and the object detection task, which makes the head location may mislead the object detection process. Thus, we propose a novel ``head-delay" strategy to resolve this problem. It first employs the backbone network to tackle the scene image $\mathbf{I}_{\rm s}$ and produces the general feature. Then, the head location cues are supplied into the gaze prediction network. In particular, we use five convolutional layers with stride 2 to tackle the head location map and combine the generated feature with the scene feature.

Based on the above two improvements, we follow \cite{chong2020detecting} to perform gaze object prediction. As shown in Fig.~\ref{figGazePrediction}, the gaze prediction network mainly consists of three modules, \ie the head module, the scene module, and the gaze prediction module. In the scene module, we first process the head location image $\mathbf{H}$ by five convolutional layers and concatenate the output with the gaze-specific scene feature $\mathbf{f}_{\rm gaze}^{\rm s}$ to make the feature map aware of the gaze position. 
Then, in the head module, we process the head location image $\mathbf{H}$ by three max-pooling with a flatten operation, then concatenate the output with the gaze-specific head feature $\mathbf{f}_{\rm gaze}^{\rm s}$ together to generate an attention map. Finally, we feed the element-wise product of this attention map and the output of the scene module into the gaze prediction module to get the gaze heatmap $\mathbf{M}$. Particularly, the gaze prediction module is an encoder-decoder structure,  the encoder consists of two convolutional layers and the decoder consists of three deconvolutional layers.

\subsection{Object detection}
\label{secObjectDetection}
\begin{figure}[!t]
\centering
\setlength{\abovecaptionskip}{0.1cm}
\includegraphics[width=0.9\linewidth]{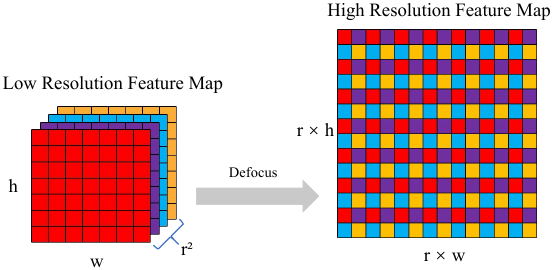}
\caption{Illustration of the \textit{Defocus} operation. One channel in the high-resolution feature map is transformed from $r^{2}$ channels in the low-resolution feature map.}
\vspace{-0.4cm}
\label{pixelshuffle}
\end{figure}

As shown in Fig.~\ref{figGOPFramework}, in the object detection branch, we firstly propose a new layer named \textit{Defocus} to generate object-specific feature $\mathbf{f}_{\rm det}^{\rm s}$, then we adopts the detection head of YOLOv4 \cite{bochkovskiy2020yolov4} to precisely discover objects in the retail scenario. 

Given an image, the backbone network tackles the entire scene image and outputs general features $\mathbf{f}_{\rm g}^{\rm s}$. An intuitive strategy to precisely detect multiple small objects in the retail scenario is to enlarge the scene image to generate high-resolution features. However, the enlarged scene images contribute little to the gaze prediction process and increase the computational burdens. Another strategy is to interpolate the feature map into a high resolution, which would lose some valuable information or bring extra computational costs. For example, considering a feature map $\mathbf{x} \in \mathbb{R}^{2048 \times 7 \times 7}$, it contains 2048-D features and exhibits spatial size $7 \times 7$. If we first compress the channel dimension via a convolutional layer and then interpolate the feature map to extend the dimension, some valuable information would be lost in the channel compressing process. Alternatively, if we directly interpolate the feature map by a factor of 2, the output would be $\mathbf{x}' \in \mathbb{R}^{2048 \times 14 \times 14}$, which increases computational costs for subsequent steps.

To resolve the aforementioned problem, we develop the \textit{Defocus} layer to enlarge the feature map without losing information or taking extra computations. This \textit{Defocus} layer is an reverse operation to the \textit{focus} layer \cite{YOLOv5}.
As shown in Fig.~\ref{pixelshuffle}, given an enlarging ratio $r$, we first rearrange elements in the feature map and shrink the channel dimension by a factor of $1/r^{2}$, then zoom the height and width by a factor of $r$. This rearranging operation requires fewer computational resources, and can hold all information in the produced object-specific feature $\mathbf{f}_{\rm det}^{\rm s}$.

After obtain the object-specific feature $\mathbf{f}_{\rm det}^{\rm s}$, the detection network uses the feature pyramid structure to fuse features from different blocks, where feature maps are interpolated to pursue a consistent spatial size. We also replace the interpolation process with the proposed \textit{Defocus} layer, which requires fewer computational sources but achieves precise detection results. Moreover, we find one high-resolution feature map is sufficient to detect retail objects and remove other detection heads, which saves 49.8\% computational costs, as multiple objects in the retail scene do not exhibit drastic variations with respect to the object size.

\subsection{Energy aggregation loss}
\label{Box energy}

The studied gaze object prediction task usually requires high-quality gaze heatmaps to generate precise results. Take the model of \cite{chong2020detecting} as an example, it achieves an AUC of 0.952 while the L$_2$ distance and angular error are 0.075 and 15.1$^\circ$, respectively. Although numerical values of these two errors seem minor, such errors would cause apparent departure on the gaze point and lead to incorrect gaze object predictions. A high-quality gaze heatmap should accurately hit the gaze point, so its energy should be aggregated on the ground truth stared object. Thus, we propose the energy aggregation loss and utilize the ground truth to guide the gaze estimation process.
\begin{figure}[!t]
\centering
\setlength{\abovecaptionskip}{0.1cm}
\includegraphics[width=0.85\linewidth]{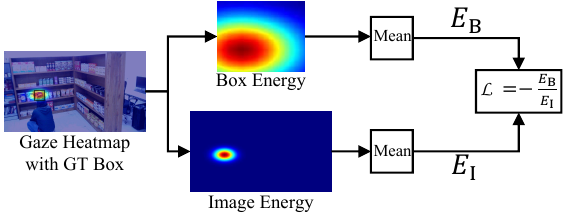}
\caption{Calculation procedure of the energy aggregation loss. ``Mean" denotes the element-wise mean operation.}
\vspace{-0.5cm}
\label{figTask}
\end{figure}

 As shown in Fig.~\ref{figTask}, given a heatmap $\mathbf{M}$ predicted by the gaze prediction module, the energy of each pixel can be represented as $\mathbf{M}_{i,j}$. If the ground truth gaze object box is defined as $\mathbf{b} = (x_{1}, y_{1}, x_{2}, y_{2})$ where $ (x_{1}, y_{1}) $ and $ (x_{2}, y_{2}) $ are coordinates of the top left and bottom right corners, we can get the average energy within this box by:
\begin{equation}
    E_{b} = \frac{1}{N} \sum^{x_{2}}_{i=x_{1}} \sum^{y_{2}}_{j=y_{1}} \mathbf{M}_{i,j},
    \label{eqMeanEng}
\end{equation}
Where $N$ indicates the number of elements within box $\mathbf{b}$. In addition, we can calculate the total energy within the image $E_{\rm I}$ by calculate the average value of all spatial bins. To guide the heatmap aggregated on the stared object box, we define the energy aggregation loss as:
\begin{equation}
    \mathcal{L}_{eng} = - \ \frac{E_{\rm b}}{E_{\rm I}}.
\end{equation}

\subsection{Training and inference}

\noindent
\textbf{Training.} 
For object detection, we assign three default boxes at each spatial location, their spatial sizes are $12 \times 16$, $19 \times 40$, and $28 \times 64$, respectively.
As our detection branch is based on the detection head of YOLOv4~\cite{bochkovskiy2020yolov4}, we employ the same method as YOLOv4 to calculate detection loss $\mathcal{L}_{\rm det}$, which jointly considers the detection confidence score, category classification score, and bounding box regression. In the gaze prediction branch, we first generate the ground truth heatmap from the ground truth gaze point by applying Gaussian blur and then calculate the heatmap loss $\mathcal{L}_{\rm gaze}$ via measuring the MSE loss between the predicted heatmap and the ground truth heatmap. In addition, the energy aggregation loss $\mathcal{L}_{\rm eng}$ uses the ground truth object box to guide the training of the gaze prediction branch. The total loss used in the training process consists of these three terms:
\begin{equation}
    \mathcal{L}_{\rm total} = \mathcal{L}_{\rm det}+ \mathcal{L}_{\rm gaze} + \mathcal{L}_{\rm eng}.
\end{equation}

\begin{figure}[!t]
\centering
\setlength{\abovecaptionskip}{0.1cm}
\includegraphics[width=0.8\linewidth]{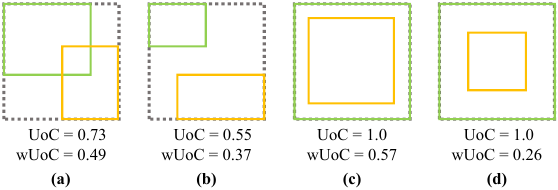}
\caption{Illustration of our proposed metric. In cases (a) and (b), the wUoC metric gets small with the decrease of the overlapping area. In cases (c) and (d), wUoC can reveal the similarity when one box fully covers the other.}
\vspace{-0.5cm}
\label{figWUoC}
\end{figure}

\noindent
\textbf{Inference.} In the inference process, the entire image and cropped head image are first processed by the proposed specific-general-specific mechanism to generate task-specific features. Then the gaze prediction branch estimates the gaze heatmap and the object detection branch predicts the object bounding boxes. Afterward, we follow Eq.~\ref{eqMeanEng} to calculate the mean energy of each box and select the box with the max mean energy as the predicted gaze object.

\subsection{Evaluation metric}
\label{secMetric}
As a newly proposed task, our studied gaze object prediction requires a proper evaluation metric. The object detection community first measures the intersection over union (IoU) between predicted and ground truth bounding boxes, then calculates the average precision (AP) to measure the performance. The AP score would be zero if there exists no overlapping area (\ie IoU=0) and make AP cannot reveal the distance between prediction and ground truth. However, such distance is meaningful when performing gaze object prediction in the retail scenario. For example, the retail system first predicts the stared object and then sequentially recommends this object and its neighboring ones to the customer.

A reasonable solution is to make the traditional IoU metric aware of bounding box distances to alleviate this issue. Given the prediction box $\mathbf{p}$ and the ground truth box $\mathbf{g}$, we calculate their minimum closure and obtain a bounding box $\mathbf{a}$. As shown in Fig.~\ref{figWUoC} (a), the union over closure $\text{UoC} = \frac{\mathbf{p} \cup \mathbf{g}}{\mathbf{a}}$ can reveal the distance between prediction and ground truth. As shown in Fig.~\ref{figWUoC} (b), even if there is no overlapping, the UoC metric can reveal the distance as well. However, when one box fully covers the other, the UoC metric would become 1, and cannot distinguish cases like Fig.~\ref{figWUoC} (c) and Fig.~\ref{figWUoC} (d).

We further introduce a size similarity weight into the UoC metric. The size similarity weight considers the area of two boxes and can be defined as $ w = \min{(\frac{\mathbf{p}}{\mathbf{g}}, \frac{\mathbf{g}}{\mathbf{p}})}$. Thus, our proposed metric can be formulated as:
\begin{equation}
    \text{wUoC} = \min{(\frac{\mathbf{p}}{\mathbf{g}}, \frac{\mathbf{g}}{\mathbf{p}})} \times \frac{\mathbf{p} \cup \mathbf{g}}{\mathbf{a}}.
    \label{wUoC}
\end{equation}
As shown in Fig.~\ref{figWUoC} (c) and Fig.~\ref{figWUoC} (d), the wUoC metric can tackle the overlapping case and adaptively reveal the prediction quality.

\begin{table}[!t]
\small
 \centering
 \setlength{\tabcolsep}{6.2pt}
 \caption{Gaze estimation performance on the GOO-Synth.}
 \vspace{-0.3cm}
\begin{tabular}{l|ccc}
    \toprule
    Setups & AUC  $\uparrow$   & Dist.  $\downarrow$ & Ang. $\downarrow$ \\
    \midrule
    Random & 0.497 & 0.454 & 77.0 \\
    Recasens \etal \cite{recasens2015they} & 0.929 & 0.162 & 33.0 \\
    Lian \etal\cite{lian2018believe} & 0.954 & 0.107 & 19.7 \\
    Chong \etal\cite{chong2020detecting} & 0.952 & 0.075 & 15.1 \\
    \midrule
    \#a Sharing backbone & 0.905 & 0.139 & 27.1 \\
    \#b Head-free scene branch & 0.941 & 0.100 & 18.8 \\
    \#c Head-delay & 0.951 & 0.091 & 16.2 \\
    \#d GaTector & \textbf{0.957} & \textbf{0.073} & \textbf{14.9} \\
    \bottomrule
    \end{tabular}%
 \label{tabCmpGaze}%
 \vspace{-0.4cm}
\end{table}%

\section{Experiments}

\subsection{Setups}

\noindent
\textbf{Dataset.} We use GOO dataset to evaluate our proposed method. GOO contains annotations of gaze point, gaze object, and bounding boxes for all objects, and have 24 different categories, where the GOO-Synth subset contains 192000 synthetic images and the GOO-Real subset contains 9552 real images. The GOO dataset exhibits the multiple challenges for the gaze object prediction task such as small size objects, multiple objects. Images in GOO dataset contains 80 objects on average.

\noindent
\textbf{Implementation details.} In the proposed GaTector, the sharing backbone network adopts ResNet-50 \cite{he2016deep}. As for object detection, we set $r=2$ in the \textit{Defocus} layer, use NMS with threshold 0.3 to remove redundant boxes, and keep the top 100 boxes for each image. As for gaze estimation, we apply Gaussian blur with kernel 3 to tackle the groundtruth gaze point. The network is optimized by the Adam \cite{kingma2014adam} algorithm for 100 epochs. We set batch size as 32, set the initial learning rate as $10^{-4}$. All experiments are implemented based on the PyTorch~\cite{NEURIPS2019_9015} and carried out on GeForce RTX 3090 GPUs.

\noindent
\textbf{Metric.} As for gaze object prediction, we use the wUoC metric as described in Section \ref{secMetric}. For gaze prediction, we follow previous works~\cite{tomas2021goo,chong2020detecting,lian2018believe} and use three metrics, \ie AUC, L2 distance (Dist.), and Angular error (Ang.). The AUC metric determines whether each cell in the spatially discretized image is classified as a gaze object. Dist. is the L2 distance between the annotated gaze location and the prediction. Ang. is the angular error of the predicted gaze. For object detection, we followed the previous works \cite{bochkovskiy2020yolov4,ren2015faster,he2017mask} to use the average precision (AP) as our metric.

\subsection{Gaze prediction}

\begin{table}[!t]
 \centering
 \small
 \setlength{\tabcolsep}{3.5pt}
  \caption{Object detection performance on the GOO-Real. The computational cost of each setup is reported in terms of Parameters (M) and FLOPs (G).}
  \vspace{-0.3cm}
    \begin{tabular}{l|cc|ccc}
    \toprule
    Setups & Para. & FLOPs & AP & AP$_{50}$ & AP$_{75}$ \\
    \midrule
    \#a YOLOv4 \cite{bochkovskiy2020yolov4} & 64.07  & 8.69  & 43.69 & 84.02 & 43.59 \\
    \#b ResNet-50 backbone & 61.65  & 7.85  & 40.87 & 82.04 & 37.48 \\
    \#c Large feature map & 40.35  & \textbf{6.17}  & 41.07 & 81.67 & 38.46 \\
    \#d Interpolation & 40.35  & 12.37  & 54.67 & 94.93 & 58.21 \\
    \#e Defocus & \textbf{39.53}  & 12.30 & \textbf{56.20} & \textbf{96.90} & \textbf{60.23} \\
    \#f GaTector & 60.78  & 18.11 & 52.25 & 91.92 & 55.34 \\
    \bottomrule
    \end{tabular}%
 \label{tabCmpDetection}%
 \vspace{-0.4cm}
\end{table}%

In traditional gaze follow works \cite{recasens2015they, lian2018believe, chong2020detecting}, two separated networks are used to tackle the scene image and the head image, respectively. Under this paradigm, Lian~\etal \cite{lian2018believe} and Chong \etal \cite{chong2020detecting} achieve promising performance, as shown in Table \ref{tabCmpGaze}. Aiming to employ a sharing backbone to perform gaze object detection, we study multiple strategies to integrate the scene and head branches.

First of all, we use ResNet-50 \cite{he2016deep} as a sharing backbone and and jointly employ the proposed SGS architecture to serve the scene and head branches. Following Chong \etal \cite{chong2020detecting}, we concatenate scene image and head location map and send the four-channel input to the scene branch. As shown in Table \ref{tabCmpGaze} setup \#a, simply sharing backbone gets an AUC of 0.905 and leads to 0.047 performance drop. We find this drop mainly comes from the random initialization of the convolutional block in the image branch. In particular, given a four-channel input, the first convolutional layer cannot use the pre-trained weights in ResNet-50 \cite{he2016deep}, and features extracted from the randomly initialized layer are insufficient to capture sensitive information.

Thus, we remove the head location map and only send the scene image to the scene branch. This experiment makes an AUC of 0.941 and verifies the necessity of proper initialization for the first convolutional layer. However, without the head location map, the scene branch still shows a performance drop concerning \cite{chong2020detecting}. Consequently, we propose the head-delay strategy, as described in Section \ref{sectionGazePrediction}. The head-delay strategy gets comparable performance with Chong's method \cite{chong2020detecting}. Moreover, the complete method of our proposed GaTector achieves high-quality gaze estimation performance and makes improvements under all three metrics.
\begin{figure*}[t]
\centering
\includegraphics[width=1\linewidth]{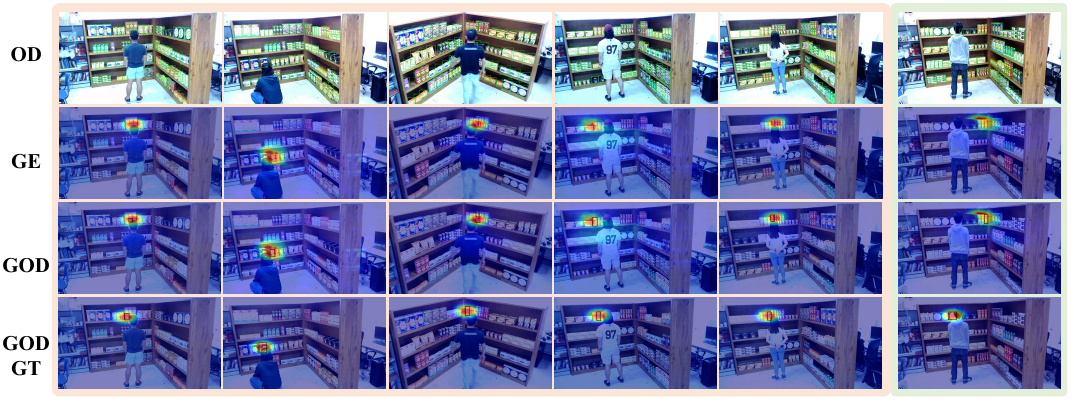}
\vspace{-0.4cm}
\caption{Visualization results of GaTector, we shows the results of object detection (OD), gaze estimation (GE), gaze object prediction (GOD), and ground truth gaze object prediction (GOD GT). We provide five successful casss (left) and one fail case (right)}
\vspace{-0.5cm}
\label{figvisualization}
\end{figure*}

\begin{table}[t]
 \centering
 \caption{Comparison of gaze object prediction performance, measured by wUoC (\%) on the GOO-Real dataset.}
 \vspace{-0.3cm}
 \small
 \setlength{\tabcolsep}{2.0pt}
    \begin{tabular}{l|ccc|cc}
    \toprule
    \multirow{2}[1]{*}{Setups} & \multirow{2}[1]{*}{Para.} & \multirow{2}[1]{*}{FLOPs} & \multirow{2}[1]{*}{AP$_{50}$} & \multicolumn{2}{c}{GOP} \\
          &       &       &     & GT gaze & Pred. gaze \\
    \midrule
    Faster-RCNN \cite{ren2015faster} & 102.70  & 32.92  & 25.47  & 12.04  & 0.69  \\
    PAA \cite{paa-eccv2020} & 93.39  & 19.40  & 54.72  & 13.61  & 1.01  \\
    RetinaNet \cite{lin2017focal} & 98.05  & 20.02  & 72.86  & 15.06  & 1.77  \\
    Sabl \cite{Wang_2020_ECCV} & 97.64  & 19.58  & 73.10  & 15.14  & 1.21  \\
    FCOS \cite{tian2019fcos} & 93.34  & 19.16  & 74.78  & 14.17  & 0.66  \\
    YOLOv4 \cite{bochkovskiy2020yolov4} & 125.52  & 18.14  & 84.02  & 16.05  & 1.56  \\
    \midrule
    GaTector & \textbf{60.78} & \textbf{18.11} & \textbf{91.92} & \textbf{20.35} & \textbf{3.31} \\
    \bottomrule
    \end{tabular}%
\vspace{-0.5cm}
 \label{tab:addlabel}%
\end{table}%

\subsection{Object detection}
Table \ref{tabCmpDetection} studies the object detection performance under different setups. First of all, vanilla YOLOv4 \cite{bochkovskiy2020yolov4} with CSPDarknet53 initialization gets an AP of 43.69\%, which requires 64.07M parameters. In the proposed GaTector, the object detection branch and the gaze prediction branch require the same backbone network. Thus, we use ResNet-50 \cite{he2016deep} to replace the CSPDarknet53 backbone and leads to a performance drop of 2.82\%. Considering the gaze object prediction dataset contains small objects with consistent size, we believe a large feature map is beneficial to detect gaze objects. Given three feature maps with different spatial sizes from YOLOv4, we perform the object detection only on the large feature map and observe slight performance improvements, as shown in Table \ref{tabCmpDetection} setup \#c. 

The small and dense objects inspire us to use large feature maps. However, due to the constraint of computational burdens and GPU memory, the spatial size of the input image is limited (\eg 224 $\times$ 224). An intuitive way to enlarge the feature map is an interpolation, which gets an AP of {54.67\%} with the cost of more parameters and increased computations. In contrast to interpolation, the proposed \textit{Defocus} layer can improve the performance to 56.2\% but requires fewer computations. As for the complete method, GaTector makes an AP of 52.25\% when only learns from the GOO-Real subset. The simplified data augmentation process primarily causes a performance gap between 56.20\% and 52.25\%. Specifically, when integrating gaze follow and object detection into a unified framework, we only keep transformations suitable to both tasks, \eg random crop and color transformation. Because GOO-Real only contains limited images, the simplified data augmentation would damage the detection performance.

\subsection{Gaze object prediction}

Table \ref{tab:addlabel} reports the gaze object detection performance under two different paradigms: separated paradigm and unified paradigm. Firstly, we can perform gaze object detection via two separated steps, \ie object detection, and gaze selection. We carry out experiments with six well-performed methods: Faster-RCNN \cite{ren2015faster}, PAA \cite{paa-eccv2020}, RetinaNet \cite{lin2017focal}, Sabl \cite{Wang_2020_ECCV} FCOS \cite{tian2019fcos} and YOLOv4 \cite{bochkovskiy2020yolov4}, including both anchor-based methods \cite{bochkovskiy2020yolov4, ren2015faster, paa-eccv2020, lin2017focal, Wang_2020_ECCV} and anchor-free methods \cite{tian2019fcos}, both one-stage methods \cite{lin2017focal, bochkovskiy2020yolov4, tian2019fcos,paa-eccv2020} and two-stage methods \cite{ren2015faster,Wang_2020_ECCV}. Trained on the GOO-Real dataset, YOLOv4 \cite{bochkovskiy2020yolov4} and FCOS \cite{tian2019fcos} achieve precise performance while the FLOPs are limited.

\begin{table*}[t]
\small
 \centering
 \caption{Ablation studies on GOO-Synth, we report the performance of gaze estimation (GE), object detection (OD), and gaze object prediction (GOP).}
  \vspace{-0.3cm}
  \begin{tabular}{l|ccc|ccc|c}
    \toprule
    \multirow{2}[1]{*}{Setups} & \multicolumn{3}{c|}{GE} & \multicolumn{3}{c|}{OD} & GOP  \\
          & AUC $\uparrow$   & Dist. $\downarrow$ & Ang. $\downarrow$  & AP  & AP$_{50}$ & AP$_{75}$ & wUoC \\
    \midrule
    \#a w/o input-specific blocks & 0.596 & 0.387 & 70.17 & \multicolumn{1}{l}{35.2} & 66.4 & 41.3 & 0.0 \\
    \#b w/o gaze-specific blocks & 0.952  & 0.114  & 20.76  & \multicolumn{1}{l}{54.7} & 93.2 & 60.5 & 24.7 \\
    \#c w/o object-specific Defocus & 0.951 & 0.075 & 15.20 & 40.9 & 81.6 & 38.3 & 24.9 \\
    \#d w/o $\mathcal{L}_{eng}$ & 0.946 & 0.103 & 17.20 & 55.8 & 94.4 & 61.5 & 25.2 \\
    GaTector & \textbf{0.957} & \textbf{0.073} & \textbf{14.91}  & \textbf{56.8} & \textbf{95.3} & \textbf{62.5} & \textbf{28.5} \\
    \bottomrule
    \end{tabular}%
    \vspace{-0.5cm}
 \label{tabAblaComplete}%
\end{table*}%

Afterward, we employ a complete GOO dataset to train the proposed GaTector model and predict the gaze heatmap. Given object detection boxes and predicted gaze heatmaps, we follow Eq.(\ref{eqMeanEng}) to select bounding boxes with high mean energy and report results for the gaze object prediction task. As shown in Table \ref{tab:addlabel}, although object detection results are acceptable, the gaze object prediction scores are low. We find the performance bottleneck lies in inaccurate gaze heatmaps. In contrast, if we utilize the ground truth heatmaps to select boxes from the detection results, the gaze object prediction performance would receive obvious gains.

Compared with the separated paradigm, our proposed GaTector adopts the unified paradigm, \ie employing a unified network to perform gaze follow, object detection and conduct gaze object prediction. As shown in Table \ref{tab:addlabel}, GaTector consistently improves the object detection performance from 84.02\% to 91.92\%. However, due to the small amount of data in the Goo-Real dataset, current methods exhibit limited performance on the gaze object prediction task. There are two potential reasons, the predicted gaze point is not accurate enough, or our metric wUoC is stricter than traditional detection performance mAP.

\subsection{Ablation studies}
\begin{table}[t]
\small
  \centering
  \caption{Analysis the bottleneck of GaTector on the GOO-Synth.}
    \begin{tabular}{cc|cc|c}
    \toprule
    \multicolumn{2}{c|}{Gaze Heatmap} & \multicolumn{2}{c|}{Object Box} & GOP \\
    \cmidrule{1-4}    GT    & Pred. &      GT & Pred. & wUoC \\
    \midrule
          & \checkmark &       & \checkmark & 28.50 \\
          & \checkmark & \checkmark &       & 29.81 \\
    \checkmark &       &       & \checkmark & 78.79 \\
    \bottomrule
    \end{tabular}%
  \label{tabAblBottleneck}%
\end{table}%

In GaTector, we propose an SGS feature extractor, a \textit{Defocus} layer and an energy aggregation loss to perform gaze object prediction. Table \ref{tabAblaComplete} reports ablation experiments about each component. Firstly, it would cause a dramatic performance drop by replacing two input-specific convolutional layers $\psi^{\rm s}(\cdot)$ and $\psi^{\rm h}(\cdot)$. This verifies the necessity of transposing inputs to a general space before sending them to the general backbone.
Besides, both gaze follow and object detection receive a certain performance drop when removing gaze-specific output blocks $\phi^{\rm s}(\cdot)$ and $\phi^{\rm h}(\cdot)$, which verifies the influence of the task-specific output convolutional layer as well. In addition, we replace the \textit{Defocus} layer with the traditional interpolation operation and observe performance drops. These three experiments prove the rationality and necessity of our proposed SGS feature extract mechanism. Finally, we can observe that removing the energy aggregation loss would damage the performance of gaze estimation, especially under the metric L2 distance and angular error.

Table~\ref{tabAblBottleneck} analyzes the performance bottleneck of our proposed GaTector. Using predicted gaze heatmap and object bounding boxes, we select a suitable bounding box by Eq.(\ref{eqMeanEng}) from prediction boxes within the heatmap, GaTector gets wUoC of {28.50\%}. If we use ground truth object boxes, the gaze object prediction performance can only be improved to {29.81\%}, which verifies the current approach has achieved accurate object detection performance. In contrast, if we use the ground truth heatmap to select boxes, the performance will receive huge improvements to 78.79\%, which indicates that the current performance bottleneck mainly lies in gaze estimation. This performance bottleneck analysis provides a promising direction for future research.

\subsection{Qualitative results}

Fig.~\ref{figvisualization} exhibits qualitative results for the proposed GaTector method. Faced with various retail scenarios (\eg various distances to the retail objects, different shooting views), our proposed GaTector can accurately localize bounding boxes and relatively precisely predict gaze heatmap, finally leading to accurate gaze object prediction results. The last row of Fig.~\ref{figvisualization} shows a failure case, where errors in the gaze heatmap mislead the gaze object prediction process.

\vspace{-0.2cm}
\section{Conclusion}
\vspace{-0.2cm}

We make an early exploration to build a unified framework to tackle the gaze object prediction task. To simultaneously conduct object detection and gaze estimation, we propose a novel SGS mechanism to extract two task-specific features with only one backbone, which can help to reduces model parameters and computational burdens. Then, we introduces two input-specific blocks before the shared backbone and three task-specific blocks after the shared backbone to take the specificity of inputs and tasks into consideration. To serve small object detection in retail scenarios, we propose a \textit{Defocus} layer to enlarge the feature map without losing information or bringing extra computations. Also, we design energy aggregation loss and employ the bounding box of the stared object to guide the gaze heatmap prediction. The promising gains of GaTector in all three tracks of the GOO dataset demonstrate the efficacy of our method, which would inspire related tasks, \eg multi-modality learning \cite{wang2021polo}, small object detection \cite{feng2020tcanet}.

\vspace{1mm}
\noindent \textbf{Limitations.} Since our GaTector is only verified on the benchmark dataset, it may encounter challenges when being applied to practical scenarios, \eg lacking of sufficient supervision\cite{wang2021pfwnet, yang2021background}, online decision\cite{yang2022colar}. This indicates a promising research direction for future works.

{\small
\bibliographystyle{ieee_fullname}
\bibliography{egbib}

\begin{thebibliography}{10}\itemsep=-1pt

\bibitem{bermejo2020eyeshopper}
Carlos Bermejo, Dimitris Chatzopoulos, and Pan Hui.
\newblock Eyeshopper: Estimating shoppers' gaze using cctv cameras.
\newblock In {\em ACM MM}, pages 2765--2774, 2020.

\bibitem{bochkovskiy2020yolov4}
Alexey Bochkovskiy, Chien-Yao Wang, and Hong-Yuan~Mark Liao.
\newblock Yolov4: Optimal speed and accuracy of object detection.
\newblock {\em arXiv preprint arXiv:2004.10934}, 2020.

\bibitem{brooks2005development}
Rechele Brooks and Andrew~N Meltzoff.
\newblock The development of gaze following and its relation to language.
\newblock {\em Developmental science}, 8(6):535--543, 2005.

\bibitem{chong2020detecting}
Eunji Chong, Yongxin Wang, Nataniel Ruiz, and James~M Rehg.
\newblock Detecting attended visual targets in video.
\newblock In {\em CVPR}, pages 5396--5406, 2020.

\bibitem{YOLOv5}
YOLO Contributors.
\newblock Yolov5.
\newblock \url{https://github.com/ultralytics/yolov5}, 2021.

\bibitem{elmadjian20183d}
Carlos Elmadjian, Pushkar Shukla, Antonio~Diaz Tula, and Carlos~H Morimoto.
\newblock 3d gaze estimation in the scene volume with a head-mounted eye
  tracker.
\newblock In {\em Proceedings of the Workshop on Communication by Gaze
  Interaction}, pages 1--9, 2018.

\bibitem{emery2000eyes}
Nathan~J Emery.
\newblock The eyes have it: the neuroethology, function and evolution of social
  gaze.
\newblock {\em Neuroscience \& biobehavioral reviews}, 24(6):581--604, 2000.

\bibitem{feng2020progressive}
Xiaoxu Feng, Junwei Han, Xiwen Yao, and Gong Cheng.
\newblock Progressive contextual instance refinement for weakly supervised
  object detection in remote sensing images.
\newblock {\em IEEE TGRS}, 58(11):8002--8012, 2020.

\bibitem{feng2020tcanet}
Xiaoxu Feng, Junwei Han, Xiwen Yao, and Gong Cheng.
\newblock Tcanet: Triple context-aware network for weakly supervised object
  detection in remote sensing images.
\newblock {\em IEEE TGRS}, 2020.

\bibitem{follmann2018mvtec}
Patrick Follmann, Tobias Bottger, Philipp Hartinger, Rebecca Konig, and Markus
  Ulrich.
\newblock Mvtec d2s: densely segmented supermarket dataset.
\newblock In {\em ECCV}, pages 569--585, 2018.

\bibitem{guan2020enhanced}
Jian Guan, Liming Yin, Jianguo Sun, Shuhan Qi, Xuan Wang, and Qing Liao.
\newblock Enhanced gaze following via object detection and human pose
  estimation.
\newblock In {\em ICMM}, pages 502--513. Springer, 2020.

\bibitem{harwood2014mobile}
Tracy Harwood and Martin Jones.
\newblock Mobile eye-tracking in retail research.
\newblock In {\em Current trends in eye tracking research}, pages 183--199.
  Springer, 2014.

\bibitem{he2017mask}
Kaiming He, Georgia Gkioxari, Piotr Doll{\'a}r, and Ross Girshick.
\newblock Mask r-cnn.
\newblock In {\em ICCV}, pages 2961--2969, 2017.

\bibitem{he2016deep}
Kaiming He, Xiangyu Zhang, Shaoqing Ren, and Jian Sun.
\newblock Deep residual learning for image recognition.
\newblock In {\em CVPR}, pages 770--778, 2016.

\bibitem{huang2021scribble}
Peiliang Huang, Junwei Han, Nian Liu, Jun Ren, and Dingwen Zhang.
\newblock Scribble-supervised video object segmentation.
\newblock {\em IEEE/CAA Journal of Automatica Sinica}, 9(2):339--353, 2021.

\bibitem{huang2021clrnet}
Peiliang Huang, Junwei Han, Dingwen Zhang, and Mingliang Xu.
\newblock Clrnet: Component-level refinement network for deep face parsing.
\newblock {\em IEEE TNNLS}, 2021.

\bibitem{paa-eccv2020}
Kang Kim and Hee~Seok Lee.
\newblock Probabilistic anchor assignment with iou prediction for object
  detection.
\newblock In {\em ECCV}, 2020.

\bibitem{kingma2014adam}
Diederik~P Kingma and Jimmy Ba.
\newblock Adam: A method for stochastic optimization.
\newblock {\em ICLR}, 2015.

\bibitem{land2009looking}
Michael Land and Benjamin Tatler.
\newblock {\em Looking and acting: vision and eye movements in natural
  behaviour}.
\newblock Oxford University Press, 2009.

\bibitem{li2021looking}
Yunhao Li, Wei Shen, Zhongpai Gao, Yucheng Zhu, Guangtao Zhai, and Guodong Guo.
\newblock Looking here or there? gaze following in 360-degree images.
\newblock In {\em ICCV}, pages 3742--3751, 2021.

\bibitem{lian2018believe}
Dongze Lian, Zehao Yu, and Shenghua Gao.
\newblock Believe it or not, we know what you are looking at!
\newblock In {\em ACCV}, pages 35--50. Springer, 2018.

\bibitem{lin2017focal}
Tsung-Yi Lin, Priya Goyal, Ross Girshick, Kaiming He, and Piotr Doll{\'a}r.
\newblock Focal loss for dense object detection.
\newblock In {\em ICCV}, pages 2980--2988, 2017.

\bibitem{liu2016ssd}
Wei Liu, Dragomir Anguelov, Dumitru Erhan, Christian Szegedy, Scott Reed,
  Cheng-Yang Fu, and Alexander~C Berg.
\newblock Ssd: Single shot multibox detector.
\newblock In {\em ECCV}, pages 21--37. Springer, 2016.

\bibitem{liu2021generalizing}
Yunfei Liu, Ruicong Liu, Haofei Wang, and Feng Lu.
\newblock Generalizing gaze estimation with outlier-guided collaborative
  adaptation.
\newblock In {\em ICCV}, pages 3835--3844, 2021.

\bibitem{liu2021goal}
Yang Liu, Lei Zhou, Xiao Bai, Yifei Huang, Lin Gu, Jun Zhou, and Tatsuya
  Harada.
\newblock Goal-oriented gaze estimation for zero-shot learning.
\newblock In {\em CVPR}, pages 3794--3803, 2021.

\bibitem{masse2019extended}
Benoit Mass{\'e}, St{\'e}phane Lathuili{\`e}re, Pablo Mesejo, and Radu Horaud.
\newblock Extended gaze following: Detecting objects in videos beyond the
  camera field of view.
\newblock In {\em 2019 14th IEEE International Conference on Automatic Face \&
  Gesture Recognition (FG 2019)}, pages 1--8. IEEE, 2019.

\bibitem{mukherjee2015deep}
Sankha~S Mukherjee and Neil~Martin Robertson.
\newblock Deep head pose: Gaze-direction estimation in multimodal video.
\newblock {\em IEEE TMM}, 17(11), 2015.

\bibitem{parks2015augmented}
Daniel Parks, Ali Borji, and Laurent Itti.
\newblock Augmented saliency model using automatic 3d head pose detection and
  learned gaze following in natural scenes.
\newblock {\em Vision research}, 116:113--126, 2015.

\bibitem{NEURIPS2019_9015}
Adam Paszke, Sam Gross, Francisco Massa, Adam Lerer, James Bradbury, Gregory
  Chanan, Trevor Killeen, Zeming Lin, Natalia Gimelshein, Luca Antiga, Alban
  Desmaison, Andreas Kopf, Edward Yang, Zachary DeVito, Martin Raison, Alykhan
  Tejani, Sasank Chilamkurthy, Benoit Steiner, Lu Fang, Junjie Bai, and Soumith
  Chintala.
\newblock Pytorch: An imperative style, high-performance deep learning library.
\newblock In H. Wallach, H. Larochelle, A. Beygelzimer, F. d\textquotesingle
  Alch\'{e}-Buc, E. Fox, and R. Garnett, editors, {\em NeurIPS}, pages
  8024--8035. Curran Associates, Inc., 2019.

\bibitem{recasens2015they}
Adria Recasens, Aditya Khosla, Carl Vondrick, and Antonio Torralba.
\newblock Where are they looking?
\newblock {\em NeurIPS}, 28:199--207, 2015.

\bibitem{recasens2017following}
Adria Recasens, Carl Vondrick, Aditya Khosla, and Antonio Torralba.
\newblock Following gaze in video.
\newblock In {\em ICCV}, pages 1435--1443, 2017.

\bibitem{redmon2016you}
Joseph Redmon, Santosh Divvala, Ross Girshick, and Ali Farhadi.
\newblock You only look once: Unified, real-time object detection.
\newblock In {\em CVPR}, pages 779--788, 2016.

\bibitem{redmon2017yolo9000}
Joseph Redmon and Ali Farhadi.
\newblock Yolo9000: better, faster, stronger.
\newblock In {\em CVPR}, pages 7263--7271, 2017.

\bibitem{redmon2018yolov3}
Joseph Redmon and Ali Farhadi.
\newblock Yolov3: An incremental improvement.
\newblock {\em arXiv preprint arXiv:1804.02767}, 2018.

\bibitem{ren2015faster}
Shaoqing Ren, Kaiming He, Ross Girshick, and Jian Sun.
\newblock Faster r-cnn: Towards real-time object detection with region proposal
  networks.
\newblock {\em NeurIPS}, 28:91--99, 2015.

\bibitem{saran2018human}
Akanksha Saran, Srinjoy Majumdar, Elaine~Schaertl Short, Andrea Thomaz, and
  Scott Niekum.
\newblock Human gaze following for human-robot interaction.
\newblock In {\em IROS}, pages 8615--8621. IEEE, 2018.

\bibitem{tian2019fcos}
Zhi Tian, Chunhua Shen, Hao Chen, and Tong He.
\newblock Fcos: Fully convolutional one-stage object detection.
\newblock In {\em ICCV}, pages 9627--9636, 2019.

\bibitem{tomas2021goo}
Henri Tomas, Marcus Reyes, Raimarc Dionido, Mark Ty, Jonric Mirando, Joel
  Casimiro, Rowel Atienza, and Richard Guinto.
\newblock Goo: A dataset for gaze object prediction in retail environments.
\newblock In {\em CVPR}, pages 3125--3133, 2021.

\bibitem{triesch2006gaze}
Jochen Triesch, Christof Teuscher, Gedeon~O De{\'a}k, and Eric Carlson.
\newblock Gaze following: why (not) learn it?
\newblock {\em Developmental science}, 9(2):125--147, 2006.

\bibitem{wang2021polo}
Binglu Wang, Le Yang, and Yongqiang Zhao.
\newblock Polo: Learning explicit cross-modality fusion for temporal action
  localization.
\newblock {\em IEEE Sign. Process. Letters}, 28:503--507, 2021.

\bibitem{wang2021exploring}
Binglu Wang, Xun Zhang, and Yongqiang Zhao.
\newblock Exploring sub-action granularity for weakly supervised temporal
  action localization.
\newblock {\em IEEE TCSVT}, 2021.

\bibitem{wang2021multiple}
Binglu Wang, Yongqiang Zhao, and Xuelong Li.
\newblock Multiple instance graph learning for weakly supervised remote sensing
  object detection.
\newblock {\em IEEE TGRS}, 60:1--12, 2021.

\bibitem{wang2021pfwnet}
Binglu Wang, Yongqiang Zhao, and Yani Zhang.
\newblock Pfwnet: Pretraining neural network via feature jigsaw puzzle for
  weakly-supervised temporal action localization.
\newblock {\em Neurocomputing}, 443:162--173, 2021.

\bibitem{Wang_2020_ECCV}
Jiaqi Wang, Wenwei Zhang, Yuhang Cao, Kai Chen, Jiangmiao Pang, Tao Gong,
  Jianping Shi, Chen~Change Loy, and Dahua Lin.
\newblock Side-aware boundary localization for more precise object detection.
\newblock In {\em ECCV}, 2020.

\bibitem{wang2022contextual}
Shunzhou Wang, Tianfei Zhou, Yao Lu, and Huijun Di.
\newblock Contextual transformation network for lightweight remote-sensing
  image super-resolution.
\newblock {\em IEEE TGRS}, 60:1--13, 2022.

\bibitem{wang2022detail}
Shunzhou Wang, Tianfei Zhou, Yao Lu, and Huijun Di.
\newblock Detail-preserving transformer for light field image super-resolution.
\newblock In {\em AAAI}, 2022.

\bibitem{wei2019rpc}
Xiu-Shen Wei, Quan Cui, Lei Yang, Peng Wang, and Lingqiao Liu.
\newblock Rpc: A large-scale retail product checkout dataset.
\newblock {\em arXiv preprint arXiv:1901.07249}, 2019.

\bibitem{yang2022colar}
Le Yang, Junwei Han, and Dingwen Zhang.
\newblock Colar: Effective and efficient online action detection by consulting
  exemplars.
\newblock In {\em CVPR}, 2022.

\bibitem{yang2021background}
Le Yang, Junwei Han, Tao Zhao, Tianwei Lin, Dingwen Zhang, and Jianxin Chen.
\newblock Background-click supervision for temporal action localization.
\newblock {\em IEEE TPAMI}, 2021.

\bibitem{yao2020automatic}
Xiwen Yao, Xiaoxu Feng, Junwei Han, Gong Cheng, and Lei Guo.
\newblock Automatic weakly supervised object detection from high spatial
  resolution remote sensing images via dynamic curriculum learning.
\newblock {\em IEEE TGRS}, 59(1):675--685, 2020.

\bibitem{zhao2020learning}
Hao Zhao, Ming Lu, Anbang Yao, Yurong Chen, and Li Zhang.
\newblock Learning to draw sight lines.
\newblock {\em IJCV}, 128(5):1076--1100, 2020.

\bibitem{zheng2020distance}
Zhaohui Zheng, Ping Wang, Wei Liu, Jinze Li, Rongguang Ye, and Dongwei Ren.
\newblock Distance-iou loss: Faster and better learning for bounding box
  regression.
\newblock In {\em AAAI}, volume~34, pages 12993--13000, 2020.

\bibitem{zhou2020matnet}
Tianfei Zhou, Jianwu Li, Shunzhou Wang, Ran Tao, and Jianbing Shen.
\newblock Matnet: Motion-attentive transition network for zero-shot video
  object segmentation.
\newblock {\em IEEE TIP}, 29:8326--8338, 2020.

\bibitem{zhou2022group}
Tianfei Zhou, Liulei Li, Xueyi Li, Chun-Mei Feng, Jianwu Li, and Ling Shao.
\newblock Group-wise learning for weakly supervised semantic segmentation.
\newblock {\em IEEE TIP}, 31:799--811, 2022.

\bibitem{zhou2021cascaded}
Tianfei Zhou, Siyuan Qi, Wenguan Wang, Jianbing Shen, and Song-Chun Zhu.
\newblock Cascaded parsing of human-object interaction recognition.
\newblock {\em IEEE TPAMI}, 2021.
\newblock doi:{10.1109/TPAMI.2021.3049156}.

\bibitem{zhou2020cascaded}
Tianfei Zhou, Wenguan Wang, Siyuan Qi, Haibin Ling, and Jianbing Shen.
\newblock Cascaded human-object interaction recognition.
\newblock In {\em CVPR}, pages 4263--4272, 2020.

\bibitem{zhu2012face}
Xiangxin Zhu and Deva Ramanan.
\newblock Face detection, pose estimation, and landmark localization in the
  wild.
\newblock In {\em CVPR}, pages 2879--2886. IEEE, 2012.

\end{thebibliography}
}

\end{document}



\title{Supplementary Materials for \\ GaTector: A Unified Framework for Gaze Object Prediction}
\author{Binglu\ Wang\thanks{Corresponding author.}, \ \ \ \ Tao\ Hu, \ \ \ \ Baoshan\ Li, \ \ \ \ Xiaojuan\ Chen,\ \ \ \ Zhijie\ Zhang\\Xi’an University of Architecture and Technology, \ \ China}
\maketitle

\section{Process to calculate the training loss.}

The training process of our GaTector is driven by three loss terms,~\ie object detection loss $\mathcal{L}_{det}$, gaze estimation loss $\mathcal{L}_{gaze}$ and our proposed energy aggregation loss $\mathcal{L}_{eng}$. The energy aggregation loss $\mathcal{L}_{eng}$ is illustrated in SubSection 3.4 of our manuscript. Here, we elaborate detailed processes to calculate the object detection loss $\mathcal{L}_{det}$ and the gaze estimation loss $\mathcal{L}_{gaze}$. For fair comparison, we keep the same setups with YOLOv4 \cite{bochkovskiy2020yolov4} when calculating the object detection loss $\mathcal{L}_{det}$, and keep the same setups with Chong \etal~\cite{chong2020detecting} when calculating the gaze estimation loss $\mathcal{L}_{gaze}$.

\noindent
\textbf{Object detection.} Given a predicted box $(x, y, w, h, p, \mathbf{s})$, $(x,y)$ indicates the central point, $(w,h)$ indicates width and height, $p$ is the predicted overlap and $\mathbf{s} = [s_{0}, s_{1}, ..., s_{C}]$ is the predicted classification score. The corresponding ground truth can be represented as $(x^{g}, y^{g}, w^{g}, h^{g}, o, \mathbf{y})$, where $o$ indicates the ground truth overlap and $\mathbf{y} = [y_{0}, y_{1}, ..., y_{C}]$, $ y_{c} \in \{0, 1\}$ represents whether this bounding box belongs to the $c^{th}$ category. The object detection loss $\mathcal{L}_{det}$ jointly considers classification, overlapping and box regression.
\begin{equation}
    \mathcal{L}_{det} = \mathcal{L}_{det}^{cls} + \mathcal{L}_{det}^{o} + \mathcal{L}_{det}^{reg}.
\end{equation}

The classification term calculates the binary cross-entropy loss:
\begin{equation}
    \mathcal{L}_{det}^{cls} = \frac{1}{C+1} \sum_{c=0}^{C} - [y_{c}{\rm log}\hat{s}_{c} + (1 - y_{c}){\rm log}(1 - \hat{s}_{c})],
\end{equation}
where $\hat{s}_{c}$ indicates the classification score after the sigmoid activation.

The overlapping loss adopts the binary cross-entropy loss as well:
\begin{equation}
    \mathcal{L}_{det}^{o} = - [o{\rm log}\hat{p} + (1 - o){\rm log}(1 - \hat{p})],
\end{equation}
where $\hat{p}$ indicates the overlap score after the sigmoid activation.

As for box regression, we adopt the CIoU loss \cite{zheng2020distance}. Succinctly, the regression loss can be calculated as follows:
\begin{equation}
    \mathcal{L}_{det}^{reg}=1 - IoU+\frac{\rho^{2}\left((x, y), (x^{g}, y^{g})\right)}{d^{2}}+\alpha v,
\end{equation}
where $IoU$ indicates the intersection over union between the predicted box and the ground truth box. The term $\frac{\rho^{2}\left((x, y), (x^{g}, y^{g})\right)}{d^{2}}$ aims to minimize the distances between central points of two boxes, where $\rho^{2}\left((x, y), (x^{g}, y^{g})\right)$ indicates the Euclidean distance and $d$ represents the diagonal length of the smallest enclosing box covering the two boxes. In addition, the term $\alpha v$ measures the consistency of aspect ratio, where $v$ is defined as:
\begin{equation}
    v=\frac{4}{\pi^{2}}\left(\arctan \frac{w^{g}}{h^{g}}-\arctan \frac{w}{h}\right)^{2}.
\end{equation}
The coefficient $\alpha$ can be calculated as:
\begin{equation}
    \alpha=\frac{v}{(1-IoU)+v}.
\end{equation}
Please refer to \cite{zheng2020distance} for more details about CIoU loss.

\noindent
\textbf{Gaze estimation.} We follow Chong \etal~\cite{chong2020detecting} to calculate the gaze estimation loss $\mathcal{L}_{gaze}$. Given the annotated gaze point $\mathbf{q} = (q_{x}, q_{y})$, we apply the Gaussian blur to generate the vanilla ground truth heatmap $\mathbf{T}'$.
\begin{equation}
    \begin{split}
        \mathbf{T}' = \frac{1}{2 \pi \sigma_{x} \sigma_{y} } \exp \left[-\frac{1}{2}\left(\frac{\left(x-q_{x}\right)^{2}}{\sigma_{x}^{2}}+\frac{\left (y-q_{y}\right)^{2}}{\sigma_{y}^{2}}\right)\right].
    \end{split}
    \label{eqGaussian}
\end{equation}
In Eq.(\ref{eqGaussian}), $\sigma_{x}$ and $\sigma_{y}$ indicate the standard deviation. We follow Chong \etal~\cite{chong2020detecting} and set $\sigma_{x} = 3$, $\sigma_{y} = 3$. Afterwards, we normalize the headmap and obtain the ground truth heatmap $\mathbf{T} = \mathbf{T} / {\rm max}(\mathbf{T}')$. Given a predicted heatmap $\mathbf{M} \in \mathbb{R}^{H \times W}$, we calculate the mean square error to obtain the gaze estimation loss $\mathcal{L}_{gaze}$:
\begin{equation}
    \mathcal{L}_{gaze} = \frac{1}{H \times W} \sum_{i=1}^{H} \sum_{j=1}^{W} (M_{i,j} - T_{i,j})^{2}.
\end{equation}

\begin{figure*}[htbp]
\centering
\includegraphics[width=1\linewidth]{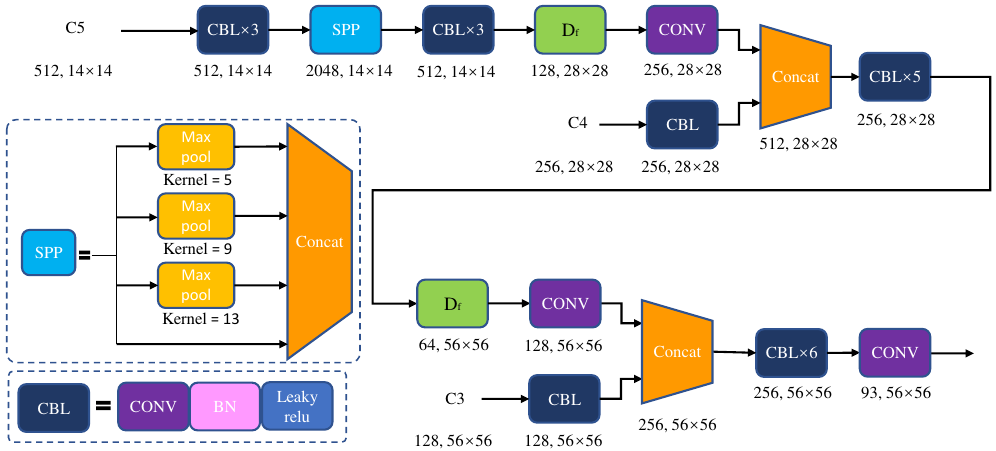}
\caption{Detailed network architecture of our object detection branch. Df, CONV, and BN indicate the defcous, convolution, and batch normalization operations, respectively. Under each operation, we present the size of the feature map in the form of (channel, height$\times$width).}
\label{figDetection}
\end{figure*}

\begin{figure}[thbp]
\centering
\includegraphics[width=0.8\linewidth]{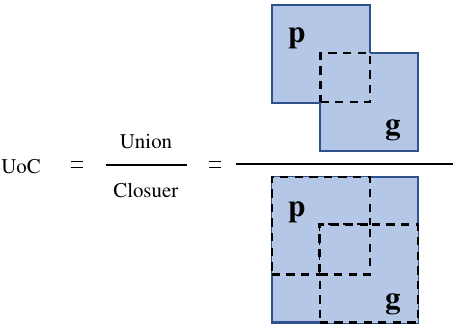}
\caption{Calculation process of union over closure.}
\label{UoC}
\end{figure}

\section{Process to calculate the wUoC metric.}
\label{wUoC_section}
In our paper, we propose the wUoC metric to measure the performance of gaze object prediction. Given the predicted box $\mathbf{p}$ and the ground truth box $\mathbf{g}$, we calculate their minimum closure and obtain a bounding box $\mathbf{a}$. Then, Figure \ref{UoC} illustrates the process to calculate the UoC (union over closure).

Afterwards, we further introduce a size similarity weight into the UoC metric. The size similarity weight considers the area of two boxes and can be defined as $\min{(\frac{\mathbf{p}}{\mathbf{g}}, \frac{\mathbf{g}}{\mathbf{p}})}$. Thus, our proposed metric can be formulated as:
\begin{equation}
    \text{wUoC} = \min{(\frac{\mathbf{p}}{\mathbf{g}}, \frac{\mathbf{g}}{\mathbf{p}})} \times \frac{\mathbf{p} \cup \mathbf{g}}{\mathbf{a}}.
    \label{wUoC}
\end{equation}

\begin{figure}[thbp]
\centering
\includegraphics[width=1\linewidth]{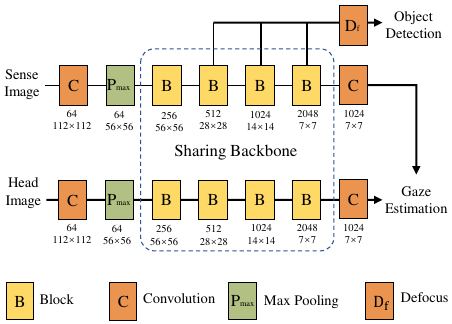}
\caption{Detailed network architecture of the specific-general-specific feature extractor. Under each operation, we present the size of the feature map in the form of (channel, height$\times$width).}
\label{figBackbone}
\end{figure}

\section{Detailed architecture of each component}

\begin{figure}[thbp]
\centering
\includegraphics[width=1\linewidth]{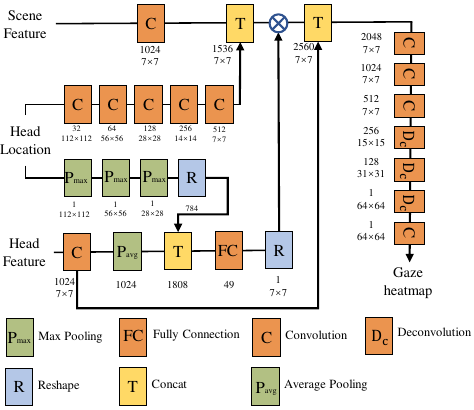}
\caption{Detailed network architecture of the gaze estimation branch. Under each operation, we present the size of the feature map in the form of (channel, height$\times$width).}
\label{figGaze}
\end{figure}

Figure \ref{figBackbone} illustrates the detailed architecture of our backbone network. Given a sense image and a head image, we first employ two specific convolutional layers to convert these two inputs into the general space. Then, a sharing backbone network with four blocks processes two inputs and generates features. Finally, we select features from the last three backbone layers~(\ie $C_{3}$, $C_{4}$, $C_{5}$), use the defcous layer to enlarge the feature map and prepare specific inputs for the object detection branch. Simultaneously, we utilize two convolutional layers to prepare inputs for the gaze estimation branch as well.

Figure \ref{figDetection} presents the detection branch. There are three inputs with different size, \ie $C_{5}$, $C_{4}$, $C_{3}$, which are gradually integrated to detect objects.

Figure \ref{figGaze} presents the detailed process to estimate the gaze heatmap. Given the head location map, we employ five convolutional layers to extract the location feature, which is concatenated with the sense feature. Simultaneously, we jointly consider head location and head feature to predict an attention map, which is used to modulate the fused feature. Afterwards, we employ three convolutional layers to abstract features, use three deconvolutional layers to enlarge the feature map, and utilize a convolutional layer to estimate the gaze heatmap.

\section{mSoC metric}
\begin{figure}[t]
\centering
\includegraphics[width=0.3\linewidth]{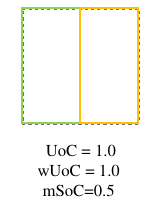}
\caption{When the area of the ground truth box is equal to the predicted box, UoC and wUoC are invalid while mSoC is valid}

\label{fig:msoc}
\end{figure}

In this paper, we use wUoC as metric for gaze object prediction. The specific details are mentioned in section \textit{Evaluation metric} of the main text, and the calculation process is mentioned in section ~\ref{wUoC_section}. However, we found that the wUoC metric is invalid when the area of the ground truth box is equal to the predicted box (as shown in Fig~\ref{fig:msoc}). Therefore, we redefine the weight of wUoC as min(p/a, g/a). We call the improved metric mSoC. The mSoC metric can be formulated as:
\begin{equation}
    \text{mSoC} = \min{(\frac{\mathbf{p}}{\mathbf{a}}, \frac{\mathbf{g}}{\mathbf{a}})} \times \frac{\mathbf{p} \cup \mathbf{g}}{\mathbf{a}}.
    \label{mSoC}
\end{equation}
Our SOTA measured by mSoC is shown in Table~\ref{sota}

\begin{table}[t]
\small
 \centering
 \setlength{\tabcolsep}{3pt}
 \caption{The performance of gaze estimation (GE), object detection (OD), and gaze object prediction (GOP, measured by mSoC).}
  \vspace{-0.3cm}
  \begin{tabular}{l|ccc|ccc|c}
    \toprule
    \multirow{2}[1]{*}{Setups} & \multicolumn{3}{c|}{GE} & \multicolumn{3}{c|}{OD} & GOP  \\
          & AUC $\uparrow$   & Dist. $\downarrow$ & Ang. $\downarrow$  & AP  & AP$_{50}$ & AP$_{75}$ & mSoC \\
    \midrule
    GaTector & \textbf{0.957} & \textbf{0.073} & \textbf{14.91}  & \textbf{56.8} & \textbf{95.3} & \textbf{62.5} & \textbf{67.94} \\
    \bottomrule
    \end{tabular}%
    \vspace{-0.5cm}
 \label{sota}%
\end{table}%

{\small
\bibliographystyle{ieee_fullname}
\bibliography{egbib}
}